\documentclass[11pt]{article}
\usepackage{arxiv}
\usepackage{booktabs}
\usepackage{amsmath}
\usepackage{amssymb}
\usepackage{graphicx}
\usepackage{url}
\usepackage[numbers]{natbib}

\title{3, 7, 14, 17, 42: Instruction Tuning Installs and Sharpens
       Number-Choice Heuristics in Language Models}
\author{Samar Ansari}
\date{\today}

\begin{document}
\maketitle

\begin{abstract}
We measure, without sampling error, the probability a language model assigns to each integer from 1 to $N$ when asked for a random number. For a fixed prompt the model defines one probability vector, which we compute by summing over each integer's scored spellings rather than estimate by sampling, removing the noise that limits earlier studies \citep{coronado2025, hopkins2023}. We evaluate 18 open-weight model lines at five ranges up to $N = 100$, under 29 decoding configurations, on two inference stacks; 15 of them have matched base and instruction-tuned checkpoints that both give usable answers. Instruction tuning makes the distribution less uniform in all 15 pairs. It installs one heuristic, round-number avoidance, and sharpens another, endpoint avoidance. Base checkpoints do not avoid round numbers once the endpoints are removed from those sets; tuned checkpoints avoid them, multiples of 10 almost completely. Base checkpoints return the endpoints 1 and $N$ at 0.30 to 0.60 of the interior rate; tuned checkpoints almost never return them. A preference for odd numbers is consistent across tuned checkpoints only at $N=100$, where base checkpoints also show it weakly. Endpoint avoidance and the odd preference fit the account \citet{kubovy1976} proposed for human number choice; round-number avoidance is not part of that account, and the midpoint avoidance it predicts is essentially absent, with one slight exception at $N=100$. Tuning does not install a new favourite integer shared across models: above $N=10$ no modal integer is shared by more than 42 percent of tuned conditions, and $P(17)$ rises in 10 of 15 pairs and falls in 5. The concentration is present in the model's own distribution: at temperature 1.0, nucleus truncation at top-$p$ 0.95 adds a median of 0.035 to the total variation distance.
\end{abstract}

\section{Introduction}
\label{sec:introduction}

Ask a chat model to pick a number between 1 and 10 and it will answer. Such
requests are common: people use models to break ties, to make arbitrary
choices and to stand in for a coin flip. The behaviour looks like a
curiosity. We treat it as a question about calibration. A model with no
reason to prefer any option should spread its probability across the
options. The same requirement arises whenever a model must make an
arbitrary choice among plausible candidates. What a model does when nothing
favours any answer is therefore a measurable property, not a parlour trick.

Existing evidence on this behaviour comes almost entirely from sampling. A
model is prompted many times under a fixed configuration and the answers
are counted \citep{hopkins2023, zhao2026}; budgets are typically
on the order of 100 calls per configuration \citep{coronado2025}. Sampling
answers a coarse question well, but it cannot answer the mechanism
question. A histogram of observed answers conflates two objects: the
distribution the model itself assigns to candidates, and whatever the
decoder does to that distribution. If instruction tuning concentrates the
distribution and truncation flattens it, or the other way round, samples
alone cannot tell which. Sampling error also limits resolution. Fine
structure, such as the mass on one integer relative to its neighbours, is
largely invisible at these budgets.

Our approach removes both obstacles for the primary measurement. For a
fixed prompt and no conversation history, a language model defines exactly
one probability distribution over continuations. Each candidate integer
corresponds to a set of token sequences; each sequence has a
computable conditional probability, and stopping after it carries its own
probability. We score the bare and leading-space spelling of every
candidate integer and sum. The result is the model's answer distribution
over those spellings, with zero sampling error. Because the quantity is
computed rather than estimated, questions of sample size and estimator
bias do not arise for the primary measurement; they apply only to the
sampling check we use for validation. We compute these distributions
for 18 open-weight model lines, at five ranges from $N=5$ to $N=100$, across 29
decoding configurations, on two inference stacks.

First, instruction tuning concentrates the distribution. In all 15
checkpoint pairs where both variants give usable answers, the tuned model sat
further from uniform than its base counterpart. At $N=30$, gemma-3-4b moves
from a total variation distance of 0.135 to 0.966. When the assistant turn
is prefilled with ``Answer:~'', the tuned checkpoint places probability
1.000 on the integer 17. Base checkpoints can be nearly uniform: OLMo-2-7B
base sits at a distance of 0.063. Second, tuned models avoid round
numbers and base models do not. With endpoints excluded from the sets, base
checkpoints place 0.92 to 1.12 of the uniform mass on interior multiples of
10, which is parity; tuned checkpoints place 0.00 to 0.12. Third, tuned
models abandon the endpoints. Relative to the average interior integer,
base models already return 1 or $N$ only 0.30 to 0.60 times as often;
tuned models return them at relative rates of 0.000 to 0.001 at every range.
Fourth, tuned models lean toward odd numbers, but the lean is consistent
across models only at $N=100$. With endpoints, multiples of 5 and the
modal integer removed, odd integers receive 2.09 times the mass of even ones
at the median, and 27 of 33 tuned conditions prefer odd. At $N=20$ and
$N=30$ the medians are 1.64 and 1.24, but a third of tuned conditions
prefer even. Base checkpoints sit near parity at $N=20$ and $N=30$ and
slightly above it at $N=100$. Endpoint avoidance and the odd preference fit
the account Kubovy and Psotka proposed for human number choice
\citep{kubovy1976}; round-number avoidance is not part of that account. The
specific integers models favour match human data at one range only: 7 at $N=10$.

Our contributions are:
\begin{itemize}
\item an enumeration method for answer distributions, replacing
  sampling with a sum over token sequences, checked against 5,000
  sampled generations for one condition and replicated on a second
  hardware stack;
\item a controlled base versus instruct comparison across 15 checkpoint
  pairs, with the answer position, decoding and hardware held fixed;
\item evidence that round-number avoidance appears only after instruction
  tuning, that endpoint avoidance already present in base checkpoints is
  pushed by tuning to near-total avoidance, and that an odd-number
  preference is consistent only at the largest range;
\item a public dataset of exact distributions, covering 85 completed runs
  and 12,325 result rows.
\end{itemize}

We claim novelty for the measurement and for the decomposition of behaviour
into model and decoder contributions, not for the phenomenon itself: that
base models are closer to uniform than tuned ones has already been reported
from sampled data \citep{westpotts2025}.

\section{Related Work}

\subsection{Human number choice}

The psychology of human random number choice has a long record. \citet{kubovy1976} asked 558 passersby for ``the first number that comes to mind between 0 and 9''. Pooled with three earlier studies, 28.4 percent of 1,770 respondents chose 7. Asked for a number between 20 and 29, 27.7 percent of 166 respondents chose 27; asked for a number between 70 and 79, only 15.5 percent of 139 chose 77, which the authors attribute to 77 looking trite and obvious. Kubovy and Psotka argue that the choice is deliberate: people try to give an answer that looks spontaneous. They suggest, explicitly as a conjecture, why 7 qualifies. It is the ``oddest'' digit: it is not in the group 2, 4, 6, 8, nor in the group 3, 6, 9; in their words, ``one can rule out 0 and 1 for being endpoints, and perhaps 5 for being a traditional midpoint'' (p.~294). Their account therefore contains endpoint avoidance, avoidance of even numbers and multiples of 3, and midpoint avoidance. It does not contain round-number avoidance. We test each part of this account against the models. \citet{towse2014} studied repeated random sequence generation over the response sets 1 to 6 and 1 to 10 and reported a small-number bias in adults. That paradigm measures sequences rather than single choices. It offers no matched baseline for single choices over larger ranges, so it cannot say what a uniform choice over 1 to 100 would look like. Our design supplies that baseline by construction, since every model distribution is compared against the exact uniform.

\subsection{Language models as random number generators}

\citet{hopkins2023} established that models asked for random numbers do not produce uniform output and cannot reproduce a requested distribution. \citet{coronado2025} is the closest prior work: six models, the ranges 1 to 5, 1 to 10 and 1 to 100, seven languages, six temperatures, and 100 sampled calls per configuration, 75,600 calls in all. The study reports modal answers of 3 and 4 at 1 to 5, 5 and 7 at 1 to 10, and 37, 47 and 73 at 1 to 100. Our instruction-tuned checkpoints reproduce these modes at the two smaller
shared ranges: 3 is the modal answer at $N=5$ in 30 of 33 model-conditions
and 7 at $N=10$ in 28 of 33. At $N=100$ the agreement is only partial: 73 is the modal
answer in 6 of 33 instruct conditions, 47 in 2 and 37 in 1, but none of the
three leads overall: the most frequent modal
integer is 42, which their study does not report. \citet{westpotts2025} report that base models are closer to uniform than aligned models and that aligned models over-produce 7; our paired design confirms the first part, since all 15 usable pairs become less uniform after instruction tuning. More broadly, \citet{kirk2024} find that RLHF reduces the diversity of model outputs. \citet{zhao2026} test 11 models against 15 target distributions and show that batch and independent-call protocols give different results. We complement this line of work by holding the prompt fixed and measuring the entire distribution rather than a sample of its draws.

\subsection{What is missing}

To our knowledge, every prior study of this question samples. Scoring candidate answers by their log-probability is standard practice in multiple-choice evaluation \citep{gao2023harness}; what we add is applying it to an open-ended choice, with a stop factor that makes the candidates mutually exclusive. Sampling yields frequencies, not probabilities, and it leaves three quantities unmeasured. We know of no earlier work that reports the probability that a model emits a valid bare integer at all, which is the precondition for every other measurement. In our data the retained instruction-tuned conditions range from 0.211 to
near 1.000 at $N=30$. Below the set-analysis threshold of 0.10 they reach 0.002, and one
condition, Mistral-7B-Instruct with prefill, falls below $10^{-4}$ at every range. We know of no earlier work that separates the model's own distribution from the decoder's effect on it, so reported non-uniformity could in principle belong to the sampler; we show it belongs to the model, with nucleus truncation at top-$p$ 0.95 adding a median of only 0.035 to total variation distance at $N=30$ and temperature 1.0. Nor do we know of a study that compares base and instruction-tuned checkpoints of the same model line with the answer position held fixed by prefill. Enumeration, which costs little more than sampling at these ranges, supplies all three.

\section{Method}
\label{sec:method}

\subsection{Enumeration}

For a fixed prompt, a language model defines a single probability vector over continuations. We compute the probability it assigns to each integer $n \in \{1,\dots,N\}$ by enumeration, rather than estimating it by sampling. The estimand is
\begin{equation}
P(n) \;=\; \sum_{v \in \mathcal{T}(n)} \Bigg[ \prod_{i} P(v_i \mid \text{prompt}, v_{<i}) \Bigg] \times P(\text{stop} \mid \text{prompt}, v),
\label{eq:estimand}
\end{equation}
where $\mathcal{T}(n)$ is the set of token sequences scored for $n$, the product runs over the tokens of sequence $v$, and the final factor is the probability that the model stops immediately after emitting $v$. The stop factor is essential. Without it, the events ``the model wrote 1'' and ``the model wrote 17'' are not mutually exclusive, because the sequence for 1 is a prefix of the sequence for 17. The resulting quantities cannot be compared across integers. Including the stop factor makes each $P(n)$ the probability of a complete, self-contained answer.

The stop factor is the total probability of one fixed set of stop tokens, the same set in every condition: the tokeniser's end-of-sequence token, the chat-template end markers that exist in the tokeniser's vocabulary (for example \texttt{<|im\_end|>}, \texttt{<|eot\_id|>}, \texttt{<|end|>} and \texttt{<|endoftext|>}), and the tokens for a newline and a double newline. The newline tokens matter most in the completion condition below, where an answer line naturally ends with a newline; the end markers matter most in the chat conditions. Counting a newline as a stop can credit a chat model that would have continued its turn on the next line; this can raise valid\_mass for chat conditions, and it can shift $P_{\mathrm{cond}}$ where the newline probability differs between integers. We did not run a check without the newline tokens.

Within each sequence we score the conditional probability of every token from the model's own next-token distributions, obtained in a single forward pass over the prompt plus the candidate. We do not reuse a key-value cache across candidates; the prompt is roughly thirty tokens, so re-running it costs almost nothing and removes any risk of one candidate leaking into the conditioning of the next.

In most byte-pair-encoding tokenisers the strings ``17'' and ``\,17'' are different token identifiers with very different probabilities. Which one the model prefers depends on whether the prompt ends at a turn boundary or mid-sentence. We therefore score both the bare form and the leading-space form of every integer, each in the tokeniser's own encoding, and sum the two probabilities. Scoring only one gives a plausible but wrong answer, and nothing in the output flags the error. A tokeniser can also spell a number in non-standard ways, for example ``1'' followed by ``7'' where ``17'' is a single token. We do not score these spellings, so $P(n)$ is exact for the scored spellings and a lower bound on the model's full probability of writing $n$. We have not measured the omitted mass; it is listed among the limitations.

Two quantities are reported for every condition. The first is \emph{valid\_mass}, defined as
\begin{equation}
\text{valid\_mass} \;=\; \sum_{n=1}^{N} P(n),
\end{equation}
the total probability that the model emits a bare valid integer and then stops. We know of no earlier report of this quantity in this literature; it measures whether the model is following the instruction at all, before any question about the distribution over integers arises. The second is the conditional distribution
\begin{equation}
P_{\mathrm{cond}}(n) \;=\; P(n) \,/\, \text{valid\_mass},
\end{equation}
which is the distribution over integers given that a valid answer was produced, and the quantity we compare against the uniform distribution.

The cost is modest. Each candidate spelling requires one short forward pass, which comes to at most 200 passes at $N{=}100$ once both spellings are counted. Prior work estimates the same distribution from around 100 sampled
generations per configuration \citep{coronado2025}. Our quantities are
computed rather than estimated, so they carry no sampling error.

\subsection{Decoding transforms}

Temperature rescaling applies to the logits at every position, so the temperature-$T$ probability of an integer is not the softmax of its temperature-1 log-probability divided by $T$. The correct quantity requires recomputing the per-position log-softmax under each temperature and each truncation rule. Our implementation does this from the stored logits of a single forward pass: for each decoding configuration we rescale the logits, apply top-$k$ and then nucleus truncation \citep{holtzman2020}, and renormalise, at every position along each candidate sequence. The full grid of 29 decoding configurations therefore costs no additional
model inference. Seven temperatures (0.1, 0.3, 0.5, 0.7, 0.8, 1.0 and 2.0)
are each combined with four truncation settings: top-$p$ at 1.0, 0.95 and
0.9 with top-$k$ disabled, and top-$p$ at 1.0 with top-$k$ set to 40. With
greedy decoding this gives $7 \times 4 + 1 = 29$ configurations. We call
temperature 1.0 with no truncation the \emph{model prior}; unless a
temperature is stated, results refer to it.

\subsection{Prompts and conditions}

The primary prompt, used verbatim, was:
\begin{quote}
``Give me a random integer between 1 and [N], inclusive. Reply with only the number.''
\end{quote}
We evaluated three prompting conditions. In the \emph{chat} condition the prompt is rendered through the model's own chat template with no prefill, so the model is free to open its turn however it likes. In the \emph{prefill} condition the chat template is used but the assistant turn is prefilled with the string ``Answer: '', forcing the answer to begin at the measured position. In the \emph{completion} condition, used for base checkpoints, the prompt is a two-shot completion ending in ``A:''. The two examples ask for an integer between 1 and 7 and between 1 and 12, and their answers are 4 and 9. Neither example range equals a test range, and neither example answer is the base modal answer at any test range.

There is an asymmetry here that must be stated plainly. The completion prompt already forces the answer position, because the model is continuing a pattern in which a number comes next. Base checkpoints under this condition are therefore compared against the instruct \emph{prefill} condition, not the no-prefill chat condition. Comparing base completions against unprefilled chat would confound the effect of post-training with the effect of forcing the answer position, and we avoid that confound by construction.

Hybrid reasoning models require a separate note. The Qwen3 line emits a
reasoning block before answering. With the default chat template the first
generated token is \texttt{<think>}, which for Qwen3-0.6B carries probability
0.99965. At the measured position the probability of a bare digit is
therefore effectively zero, and valid\_mass falls to around
$6 \times 10^{-15}$. We close the reasoning block in the template and report
the thinking-off condition throughout. This is a non-default setting, applied
uniformly; results for Qwen3 may not carry over to its default thinking mode.

We use vendor checkpoint names throughout, which means the suffix
convention differs by vendor. Google marks instruction-tuned checkpoints
\texttt{-it} and pretrained ones \texttt{-pt}; Qwen marks base
checkpoints \texttt{-Base} and leaves the instruction-tuned name
unsuffixed. So gemma-3-4b-it and Qwen3-4B are both instruction tuned,
and gemma-3-4b-pt and Qwen3-4B-Base are their base counterparts. Where a
name appears without a suffix and without \texttt{-Base}, it refers to
the pair as a whole. Table~\ref{tab:models} in Appendix~\ref{app:models}
lists every checkpoint, the conditions run and the stack each ran on.

\subsection{Validation}

We validated the enumeration against 5,000 real sampled generations from Qwen3-0.6B in the chat condition at $N{=}30$, temperature 0.7. The enumerated valid\_mass was 0.1504 against a sampled 0.1442. The enumerated modal probability was 0.579 against a sampled 0.587, with the same modal integer, 15. The enumerated total variation distance was 0.7075 against a sampled 0.7280; sampled TV is biased upward in finite samples, which accounts for the gap. This is one model in one condition; it checks the implementation, not every tokeniser.

As an independent check of the metric code itself, we passed a cryptographic random number generator and a seeded NumPy generator through the identical analysis pipeline. Both returned TV distances near zero, 0.0040 and 0.0087 respectively at 200,000 draws, consistent with Monte Carlo noise alone. A pipeline bug would have shown up here.

\subsection{Exclusions and metrics}

Rows were excluded by a pre-specified rule. If valid\_mass falls below $10^{-4}$, the conditional distribution is a ratio of near-zero quantities and is numerically meaningless; this rule excluded 1,247 of 12,325 rows, or 10.1 percent. Every analysis of the model prior, including the paired comparison, applies a stricter threshold of valid\_mass at least 0.10, because below it the conditional distribution describes a small minority of the model's behaviour. The same three model-conditions fail this threshold at every range. Both thresholds, and the rows each one removes, are reported in Appendix~\ref{app:exclusions}.

The primary metric is total variation distance from the uniform distribution, TV $= \frac{1}{2}\sum_n |P_{\mathrm{cond}}(n) - 1/N|$. It reads directly as the share of probability mass sitting on the wrong integers: 0 is perfectly uniform, and the maximum is $1 - 1/N$. Because the maximum grows with $N$, comparisons across ranges also use TV divided by $1-1/N$. We also report the entropy deficit, $\log N$ minus the entropy of $P_{\mathrm{cond}}$, in nats.

Medians over the 16 base conditions are upper medians, the 9th of 16 sorted values; over 33 instruct conditions the median is the 17th value. Comparisons across models are descriptive. Where we count models moving in one direction we give a two-sided sign test. The test treats models as independent, which they are not where they share a vendor and training lineage, so it overstates the evidence.

Where a condition was run on both stacks, tables report the CUDA value. Of the 49 conditions retained at $N=30$, 30 come from the CUDA stack and 19 from the Metal stack, which ran seven model lines that the CUDA stack did not.

\section{Measurement Validity}
\label{sec:validity}

Where possible, results were computed twice, once on each of two machines
that differ in hardware, inference stack and numeric format: an RTX 4070 Ti
running CUDA in bfloat16, and a Mac Mini 24GB running Metal/MPS in float16.
Because the numeric formats differ, the comparison tests precision
sensitivity as well as reproducibility. Identical weights, prompts and
decoding parameters were used on both stacks, so any disagreement traces back
to arithmetic rather than to sampling or configuration.

Three facts limit this comparison. First, not every model ran on both stacks: the
CUDA stack ran 32 of the 85 runs and the Metal stack 53, so seven model lines
(Qwen2.5-7B, Qwen3-8B, OLMo-2-7B, Llama-3.1-8B, Llama-3.2-1B, Llama-3.2-3B and
Mistral-7B) were measured on Metal only. Table~\ref{tab:models} in
Appendix~\ref{app:models} gives the stack for every checkpoint. Second, on the Metal stack in float16, gemma-3-4b-pt,
gemma-3-4b-it and gemma-3-1b-pt returned no valid probabilities at any range
or setting (valid\_mass of zero). We did not diagnose the cause; gemma-3-1b-it ran normally on the same stack;
these conditions rest on the CUDA stack alone. Third, the comparison below
therefore covers only conditions that produced valid output on both stacks:
26 of the 49 conditions retained at $N=30$.

For stochastic decoding we compared 3,715 overlapping model-conditions. The
median total variation distance between the two machines' distributions was
0.0167, with a 90th percentile of 0.0707. The two stacks selected the same
modal integer in 3,575 of 3,715 conditions, or 96.2 percent.
Table~\ref{tab:replication} breaks agreement down by temperature.

\begin{table}[t]
\centering
\caption{Agreement between the two inference stacks, which differ in hardware, framework and numeric format (bfloat16 on CUDA, float16 on Metal).}
\label{tab:replication}
\begin{tabular}{lrrrr}
\toprule
Temperature & $n$ & Median TV diff. & 90th pct. & Same mode \\
\midrule
greedy & 125 & 0.0000 & 0.0000 & 93.6\% \\
0.1 & 502 & 0.0000 & 0.1622 & 93.2\% \\
0.3 & 512 & 0.0168 & 0.0872 & 95.3\% \\
0.5 & 527 & 0.0234 & 0.0776 & 96.4\% \\
0.7 & 532 & 0.0225 & 0.0703 & 97.2\% \\
0.8 & 540 & 0.0212 & 0.0634 & 97.8\% \\
1.0 & 542 & 0.0194 & 0.0501 & 97.4\% \\
2.0 & 560 & 0.0142 & 0.0256 & 96.1\% \\
\bottomrule
\end{tabular}
\par\vspace{2pt}\footnotesize At low temperature the distribution is close to a point mass, so the stacks usually agree exactly and the median difference is zero. The failures are rare and total.
\end{table}

The median column in Table~\ref{tab:replication} should not be read as the
headline. Mode agreement is the informative statistic. It is lowest at the
lowest sampled temperature, 93.2 percent at $T=0.1$, and highest at $T=0.8$,
97.8 percent. At low temperature the distribution is close to a point mass,
so the two stacks usually agree exactly and the median difference is 0. The
90th percentile of 0.162 at $T=0.1$, against 0.063 at $T=0.8$, shows where
the disagreement actually lives. The failures are concentrated rather than
spread. When the top two candidates are nearly tied, small rounding
differences between bfloat16 and float16 decide the winner, and the total
variation distance jumps toward its maximum. Low-temperature disagreement is
therefore bimodal: exact agreement in most conditions, near-complete
disagreement in a few.

Greedy decoding behaves the same way. The two stacks agree in 117 of 125
conditions, 93.6 percent, with median and 90th-percentile differences of
exactly 0 and a maximum of 1.000. Greedy decoding is not unreliable in
general. It is reliable almost always and fails completely when it fails,
because a single tie at the top of the distribution flips the argmax.

In practice, reproducing a low-temperature model output across numeric
formats is not guaranteed, even with identical weights and prompts.

Truncation has the same effect for the same reason. Nucleus sampling turns
small probability differences near the cutoff into inclusion or exclusion
decisions, so a rounding difference that would be invisible in an untruncated
distribution can change which integers survive at all. Section~\ref{sec:concentration}
gives a worked case: Qwen3-0.6B in chat mode at $N=30$ and temperature 0.7 has
support 30 on both stacks without truncation, and support 5 on one stack
against 11 on the other once top-$p = 0.95$ is applied. At temperature 1.0 the
same condition keeps 21 integers on both stacks. Individual truncated
conditions should therefore be reported per stack, not pooled.

\section{Instruction Tuning Concentrates the Distribution}
\label{sec:concentration}

Previous comparisons of base and instruction-tuned models have relied on sampling, with the noise and gaps that sampling entails \citep{westpotts2025, coronado2025}. Because we recover the full answer distribution exactly, the comparison can be made directly. For each of the 15 pairs in which both checkpoints give usable answers we compute the total variation distance from uniform at $N=30$, with the base checkpoint under the model prior and the instruction-tuned checkpoint with the assistant turn prefilled, both at temperature 1.0. Table~\ref{tab:base_vs_instruct} lists the pairs. Two further pairs were run but cannot enter the comparison. Mistral-7B-Instruct with prefill almost never stops after a bare integer (valid\_mass below $10^{-4}$ at every range). gemma-2-2b-it has valid\_mass of 0.026 with prefill, below the 0.10 threshold; for the record, it too moves away from uniform, from 0.106 to 0.542.

\begin{table}[t]
\centering
\caption{Distance from uniform at $N=30$, base checkpoint against its instruction-tuned counterpart (prefill), at the model prior. TV is total variation distance; Ent.\ is the entropy deficit in nats (maximum $\log 30 = 3.401$). All 15 pairs move away from uniform on both measures. Change is computed from the printed TV values. Values are from the CUDA stack where both stacks ran (Section~\ref{sec:method}). The gemma-2-2b pair is excluded because its tuned checkpoint fails the valid\_mass threshold of 0.10 (Appendix~\ref{app:exclusions}).}
\label{tab:base_vs_instruct}
\resizebox{\linewidth}{!}{%
\begin{tabular}{lrrrrrrr}
\toprule
Pair (instruct name) & TV base & TV inst. & Change & Ent.\ base & Ent.\ inst. & Mode base & Mode inst. \\
\midrule
google/gemma-3-4b-it & 0.135 & 0.966 & +0.831 & 0.051 & 3.398 & 5 & 17 \\
google/gemma-3-1b-it & 0.178 & 0.964 & +0.786 & 0.096 & 3.377 & 30 & 15 \\
Qwen/Qwen3-4B & 0.196 & 0.946 & +0.750 & 0.137 & 3.294 & 15 & 17 \\
Qwen/Qwen2.5-3B-Instruct & 0.235 & 0.910 & +0.675 & 0.175 & 3.000 & 17 & 23 \\
Qwen/Qwen2.5-7B-Instruct & 0.110 & 0.784 & +0.674 & 0.038 & 1.929 & 22 & 17 \\
meta-llama/Llama-3.1-8B-Instruct & 0.107 & 0.623 & +0.516 & 0.040 & 1.532 & 25 & 14 \\
Qwen/Qwen3-1.7B & 0.445 & 0.960 & +0.515 & 0.582 & 3.353 & 17 & 15 \\
meta-llama/Llama-3.2-3B-Instruct & 0.079 & 0.561 & +0.482 & 0.026 & 0.995 & 12 & 14 \\
allenai/OLMo-2-1124-7B-Instruct & 0.063 & 0.520 & +0.457 & 0.029 & 0.806 & 19 & 17 \\
Qwen/Qwen3-8B & 0.475 & 0.924 & +0.449 & 0.669 & 3.187 & 17 & 17 \\
meta-llama/Llama-3.2-1B-Instruct & 0.276 & 0.645 & +0.369 & 0.221 & 1.206 & 15 & 17 \\
Qwen/Qwen3-0.6B & 0.225 & 0.504 & +0.279 & 0.172 & 0.911 & 17 & 25 \\
Qwen/Qwen2.5-0.5B-Instruct & 0.360 & 0.626 & +0.266 & 0.338 & 1.210 & 15 & 15 \\
Qwen/Qwen2.5-1.5B-Instruct & 0.498 & 0.739 & +0.241 & 0.670 & 1.546 & 17 & 14 \\
HuggingFaceTB/SmolLM2-1.7B-Instruct & 0.145 & 0.260 & +0.115 & 0.059 & 0.208 & 17 & 17 \\
\bottomrule
\end{tabular}}
\end{table}

All 15 pairs become less uniform after instruction tuning, with no exceptions (two-sided sign test, $p = 6.1 \times 10^{-5}$, with the independence caveat of Section~\ref{sec:method}). The increase in total variation distance ranges from 0.115 for SmolLM2-1.7B, which moves from 0.145 to 0.260, to 0.831 for gemma-3-4b, which moves from 0.135 to 0.966. Two further pairs are almost as large: gemma-3-1b rises from 0.178 to 0.964 and Qwen3-4B from 0.196 to 0.946. Instruction tuning does not nudge the distribution. In most pairs it rebuilds it.

The starting points differ widely across model lines. Table~\ref{tab:headline} shows the spread of behaviour at $N=30$. Several base checkpoints are close to uniform. OLMo-2-7B base has a total variation distance of 0.063 and places probability 0.045 on its modal integer, against 0.0333 for a uniform distribution over 30 integers. Its valid\_mass is 0.995, so this is a measured near-uniform preference rather than a failure to answer. Llama-3.2-3B base (0.079), Gemma-2-2b base (0.106) and Llama-3.1-8B base (0.107) are similar. These exact values confirm, without sampling error, the earlier sampled observation that base models sit closer to uniform \citep{westpotts2025}.

\begin{table}[t]
\centering
\caption{Ten most uniform and ten least uniform model-conditions at $N=30$, measured on the model prior (temperature 1.0, no truncation). A uniform distribution would give $p=0.033$. Ent.\ def.\ is the entropy deficit in nats; its maximum at $N=30$ is $\log 30 = 3.401$. Values are from the CUDA stack where both stacks ran (Section~\ref{sec:method}).}
\label{tab:headline}
\begin{tabular}{lrrrrr}
\toprule
Model [condition] & TV & Mode & $p$(mode) & Valid mass & Ent.\ def. \\
\midrule
allenai/OLMo-2-1124-7B [base] & 0.063 & 19 & 0.045 & 0.995 & 0.029 \\
meta-llama/Llama-3.2-3B [base] & 0.079 & 12 & 0.045 & 0.969 & 0.026 \\
google/gemma-2-2b [base] & 0.106 & 20 & 0.050 & 0.961 & 0.033 \\
meta-llama/Llama-3.1-8B [base] & 0.107 & 25 & 0.045 & 0.979 & 0.040 \\
Qwen/Qwen2.5-7B [base] & 0.110 & 22 & 0.044 & 0.974 & 0.038 \\
google/gemma-3-4b-pt [base] & 0.135 & 5 & 0.060 & 0.776 & 0.051 \\
HuggingFaceTB/SmolLM2-1.7B [base] & 0.145 & 17 & 0.057 & 0.876 & 0.059 \\
google/gemma-3-1b-pt [base] & 0.178 & 30 & 0.065 & 0.341 & 0.096 \\
Qwen/Qwen3-4B-Base [base] & 0.196 & 15 & 0.069 & 0.958 & 0.137 \\
Qwen/Qwen3-0.6B-Base [base] & 0.225 & 17 & 0.084 & 0.969 & 0.172 \\
\multicolumn{6}{c}{$\vdots$} \\
Qwen/Qwen3-1.7B [chat] & 0.921 & 12 & 0.807 & 1.000 & 2.843 \\
Qwen/Qwen3-8B [prefill] & 0.924 & 17 & 0.957 & 1.000 & 3.187 \\
Qwen/Qwen3-4B [chat] & 0.926 & 17 & 0.919 & 1.000 & 3.083 \\
Qwen/Qwen3-8B [chat] & 0.943 & 17 & 0.976 & 1.000 & 3.263 \\
Qwen/Qwen3-4B [prefill] & 0.946 & 17 & 0.980 & 1.000 & 3.294 \\
google/gemma-3-1b-it [chat] & 0.958 & 15 & 0.991 & 1.000 & 3.350 \\
Qwen/Qwen3-1.7B [prefill] & 0.960 & 15 & 0.993 & 1.000 & 3.353 \\
google/gemma-3-4b-it [chat] & 0.963 & 17 & 0.997 & 1.000 & 3.379 \\
google/gemma-3-1b-it [prefill] & 0.964 & 15 & 0.997 & 1.000 & 3.377 \\
google/gemma-3-4b-it [prefill] & 0.966 & 17 & 1.000 & 1.000 & 3.398 \\
\bottomrule
\end{tabular}
\par\vspace{2pt}\footnotesize Three rows with valid\_mass below 0.10 are excluded, because their conditional distributions describe only a small minority of the model's output. They are listed in Table~\ref{tab:excluded}.
\end{table}

 \begin{figure}[!tp]
 \centering
 \includegraphics[width=0.999\linewidth]{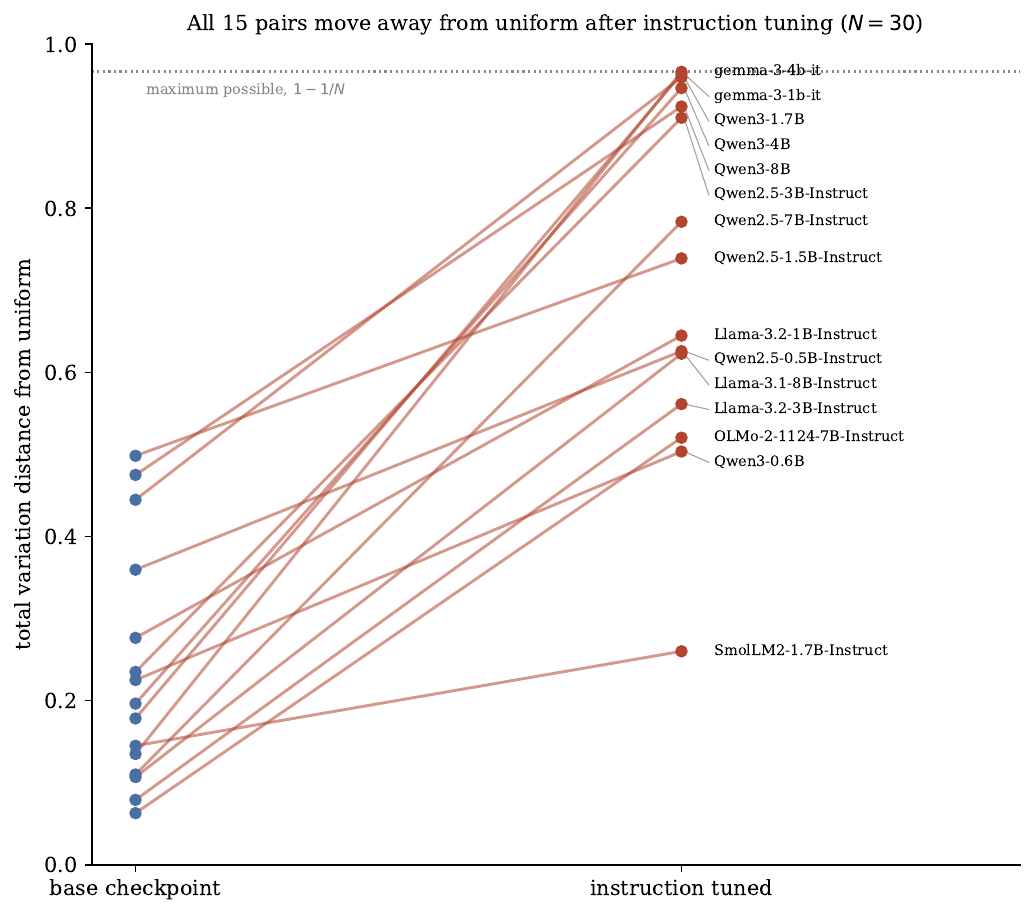}
 \caption{Total variation distance from uniform at $N=30$, base checkpoint
   against instruction-tuned counterpart, one line per pair. All 15 move
   away from uniform. The dotted line is the maximum attainable distance,
   $1-1/N$.}
 \label{fig:pairs}
 \end{figure}

\begin{figure}[!htbp]
 \centering
 \includegraphics[width=\linewidth]{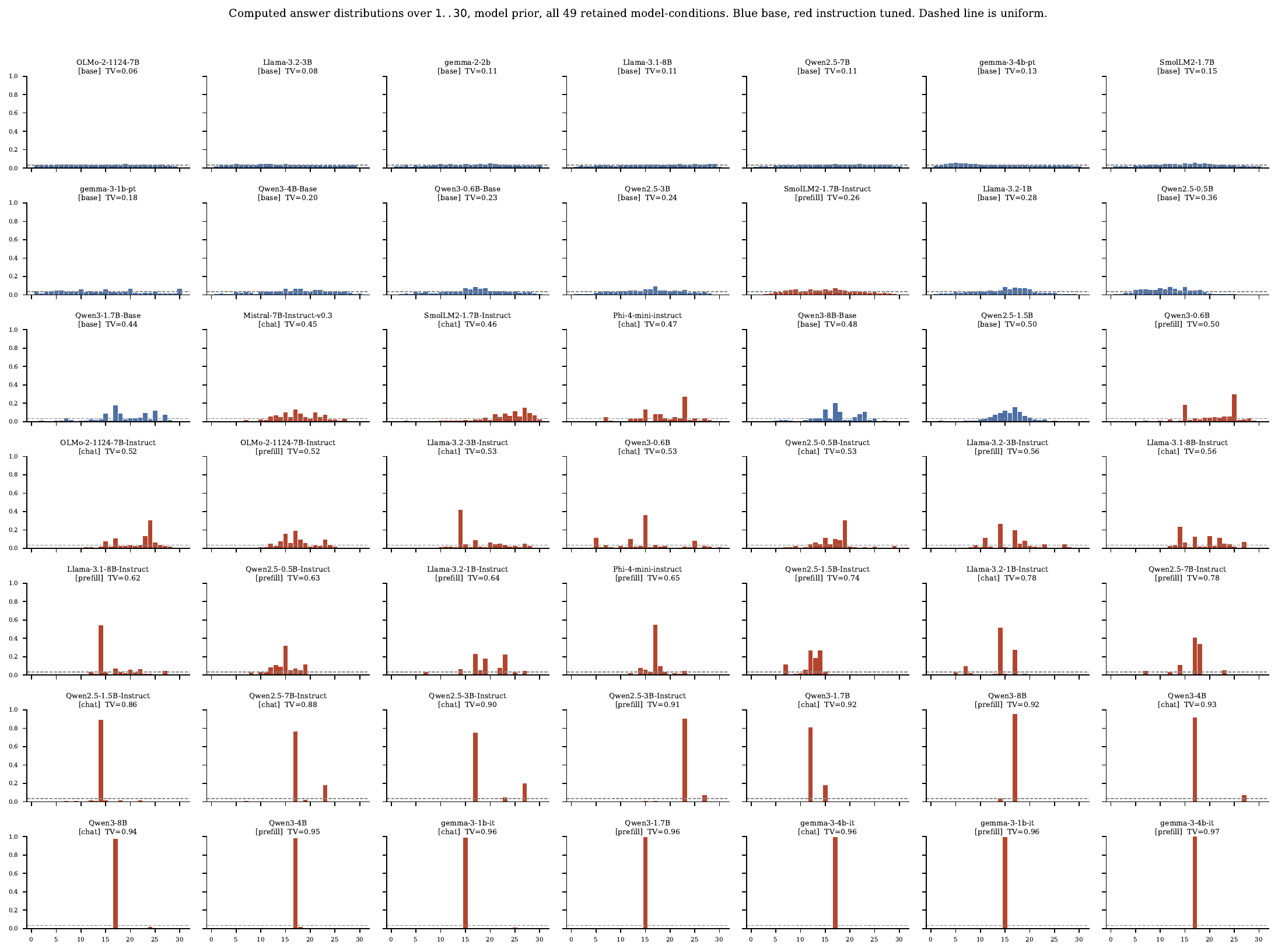}
 \caption{Computed answer distributions over $1..30$ under the model prior,
   for every retained model-condition, ordered by distance from uniform.
   Base checkpoints in blue, instruction-tuned in red. The dashed line is
   the uniform value $1/30$.}
 \label{fig:grid}
\end{figure}

At the other end, several instruct checkpoints are close to deterministic. With prefill, gemma-3-4b-it assigns probability 1.000 to 17, Qwen3-4B
assigns 0.980, and Qwen3-8B assigns 0.957. In the chat condition Qwen3-8B
reaches 0.976. The total variation ceiling for $N=30$ is $1 - 1/30 = 0.9667$, so gemma-3-4b-it, at 0.966, is within 0.001 of the largest distance any distribution over 30 integers can have from uniform. For practical purposes its answer distribution is a point mass.

The target of the concentration is not shared, and the base checkpoint does not predict it. gemma-3-4b starts with essentially no preference for 17: its base probability for 17 is 0.92 times uniform. After tuning the ratio is 29.99, an absolute probability of 0.9996. Qwen3-1.7B moves the opposite way, from 5.34 times uniform to 0.01. Its base model's most likely answer is 17 with probability 0.178, while the instruct model collapses onto 15 with probability 0.993 and all but abandons 17. Qwen2.5-1.5B also falls, from 4.63 to 0.10 times uniform. The modal integer shifts accordingly: OLMo-2-7B moves from 19 to 17, gemma-3-1b from 30 to 15, and Llama-3.1-8B from 25 to 14. Any account in which post-training installs one particular favoured number is therefore untenable. What tuning installs is concentration itself, aimed at a target that varies by model in a way the base weights do not predict.

A natural objection is that the concentration is an artefact of decoding settings rather than a property of the model. The exact computation lets us test this. We compared each distribution at $N=30$ under the model prior with its top-$p = 0.95$ counterpart at the same temperature, over all 49 retained model-conditions, counting each stack separately where a condition ran on both (75 rows). The median total variation added by truncation is 0.035 (interquartile range 0.020 to 0.048), the largest is 0.082, and the median change in modal probability is a factor of 1.04. All of these are small against the tuning effects reported above: the concentration is present in the model's own distribution before any truncation.

At temperature 0.7 the median is the same, 0.035, but one condition stands out, and it carries a caveat worth stating. Qwen3-0.6B in chat mode at temperature 0.7 is the largest single sampler effect we observe. On the CUDA stack its modal probability rises from 0.579 to 0.799 under top-$p = 0.95$ and its support falls from 30 integers to 5; on the Metal stack the same condition rises from 0.581 to 0.683 with support falling to 11. The untruncated distributions agree closely across the two machines, at 0.579 against 0.581, and the truncated ones do not. Nucleus sampling converts small probability differences near the cutoff into inclusion or exclusion decisions, so it amplifies exactly the precision sensitivity documented in Section~\ref{sec:validity}. At temperature 1.0 the same condition changes far less, from 0.358 to 0.378 in modal probability, and both stacks keep 21 integers. Single-condition sampler effects should therefore be read as specific to a stack and a temperature, and only the pooled medians as properties of the models.

Taken together, the paired comparisons support a narrow claim. Instruction tuning reduces, in every pair we measured, the entropy of the answer distribution (Table~\ref{tab:base_vs_instruct} gives the entropy deficit for each pair), sometimes to the point of determinism, but the integer it converges on is model-specific and not foreshadowed by the base checkpoint. This sharpens, with exact quantities, earlier evidence that aligned models are less random \citep{westpotts2025} and that RLHF reduces output diversity \citep{kirk2024}, and it helps explain why sampled studies report different modal numbers across models and ranges \citep{coronado2025, hopkins2023}.

\section{Which Integers Survive}
\label{sec:survive}

Total variation distance says how concentrated a distribution is, but not which
integers gained mass and which lost it. For a set $S$ of integers we report the
probability mass a model places on $S$ divided by the mass the uniform
distribution would place there. A ratio of 1.00 means the set holds exactly its
uniform share; values below 1.00 indicate avoidance and values above 1.00
indicate preference. All contrasts use the model prior, that is, temperature 1.0
with top-$p$ 1.0 and no top-$k$, and report medians across model-conditions. The
overall pattern, base checkpoints near uniform and tuned checkpoints
concentrated, matches earlier reports that base models sit closer to uniform
than aligned ones \citep{westpotts2025}.

\subsection{Endpoint avoidance}

The cleanest contrast in this section requires no set membership at all. We
compare the probability of returning 1, and the probability of returning $N$,
with the average probability of an interior integer. A ratio of 1.00 means the
endpoint is treated like any other value.

Base checkpoints return endpoints at roughly a third to a half of the interior
rate over the larger ranges: median $P(1)$ ratios of 0.406, 0.362, 0.321 and
0.356 at $N=10$, 20, 30 and 100, with $P(N)$ close behind at 0.318, 0.369,
0.300 and 0.413. At $N=5$, where only three interior integers exist, the
ratios are 0.600 and 0.601. Instruction tuning does not soften this avoidance;
it removes the endpoints. The median instruct $P(1)$ ratio is 0.0001 at $N=5$
and below 0.0001 at every larger range; the median instruct $P(N)$ ratio is
0.0007 at $N=5$ and likewise below 0.0001 elsewhere. These appear as 0.000
and 0.001 in Table~\ref{tab:disentangle}, which rounds to three decimal
places. Instruct checkpoints
essentially never return 1 or $N$.

Two features make this our strongest result. First, it is measured on the raw
distribution: no set is defined, nothing is excluded and nothing is
renormalised, so the confounds discussed below cannot touch it. Second, the
avoidance is roughly symmetric. That is not obvious, because 1 and $N$ are
salient for different reasons: 1 is the smallest counting number and a common
default in many tasks, whereas $N$ is the stated maximum, whose value changes
with the prompt. The base medians show no asymmetry: 0.321 against 0.300 at
$N=30$, and 0.356 against 0.413 at $N=100$. Part of \citet{kubovy1976}'s
proposed explanation for the predominance of 7 is that 0 and 1 are ruled out
as endpoints. The models share this part of the account; base checkpoints
show it partially, and after tuning they carry it to completion.

\subsection{Round-number avoidance}

\begin{table}[t]
\centering
\caption{Observed probability mass on each set of integers, divided by the mass a uniform distribution would place there. A ratio of 1.00 means no preference. The endpoints 1 and $N$ are excluded from every set except the endpoints row itself, and the two round-number sets are disjoint, so no row restates another. A dash means the set is empty at that range once endpoints are removed. The prime and odd-number rows remain uncontrolled and should not be read as effect sizes. Every multiple of 5 that is not a multiple of 10 is odd and almost every prime is odd, so those rows overlap each other and the round-number row; and because the modal integer stays in the set, a cell largely reports whether the mode happens to fall inside it. That is why instruct odd numbers read 4.40 at $N=5$, where the only interior odd integer is 3 and 3 is the modal answer in 30 of 33 conditions. See Table~\ref{tab:disentangle} for the versions with the mode removed. Values are from the CUDA stack where both stacks ran (Section~\ref{sec:method}).}
\label{tab:structure}
\begin{tabular}{lrrrrr}
\toprule
Set & $N{=}5$ & $N{=}10$ & $N{=}20$ & $N{=}30$ & $N{=}100$ \\
\midrule
\multicolumn{6}{l}{\textbf{base}} \\
\quad endpoints (1, N) & 0.72 & 0.46 & 0.33 & 0.38 & 0.44 \\
\quad multiples of 10 & -- & -- & 0.99 & 1.12 & 0.92 \\
\quad multiples of 5, not of 10 & -- & 1.41 & 1.31 & 1.33 & 1.10 \\
\quad primes & 1.44 & 1.25 & 1.03 & 1.05 & 1.03 \\
\quad odd numbers & 1.59 & 1.23 & 1.17 & 1.09 & 1.08 \\
\multicolumn{6}{l}{\textbf{instruct}} \\
\quad endpoints (1, N) & 0.00 & 0.00 & 0.00 & 0.00 & 0.00 \\
\quad multiples of 10 & -- & -- & 0.12 & 0.06 & 0.00 \\
\quad multiples of 5, not of 10 & -- & 0.24 & 0.54 & 0.59 & 0.16 \\
\quad primes & 2.20 & 2.12 & 0.97 & 1.17 & 1.33 \\
\quad odd numbers & 4.40 & 2.19 & 1.20 & 1.54 & 1.21 \\
\bottomrule
\end{tabular}
\end{table}

Table~\ref{tab:structure} applies the same ratio to two disjoint
round-number classes, the multiples of 10 and the multiples of 5 that are not
multiples of 10, and to the primes and the odd numbers. Both round-number
sets exclude the
endpoints 1 and $N$, which are measured separately above, so neither row
restates the endpoint effect under a second name. We defer the primes to the
final subsection, where a confound with the multiples of 3 is discussed.

Base checkpoints do not avoid round numbers. Interior multiples of 10 sit at
0.99 of their uniform share at $N=20$, 1.12 at $N=30$ and 0.92 at $N=100$,
which is parity within the precision of these measurements. Interior multiples
of 5 sit above parity throughout, at 1.41 at $N=10$, 1.31 at $N=20$,
1.33 at $N=30$ and 1.10 at $N=100$. On the endpoint-corrected measurement,
round-number avoidance is absent from pretraining altogether.

Instruction-tuned checkpoints avoid both classes, and avoid the rounder one
more. Multiples of 10 retain 0.12 of their uniform share at $N=20$, 0.06 at
$N=30$ and 0.00 at $N=100$. Multiples of 5 that are not multiples of 10 retain
0.24 at $N=10$, 0.54 at $N=20$, 0.59 at $N=30$ and 0.16 at $N=100$. Because
the two sets share no members, the gap between them compares two classes of
integer rather than a set against a superset of itself. At $N=30$ the
multiples of 5 retain roughly ten times the share of the multiples of 10, and
the ordering holds at every range where both are defined: avoidance
strengthens as a number becomes rounder. Across the base-to-instruct
transition at $N=30$, multiples of 10 fall from 1.12 to 0.06, a nineteen-fold
reduction, while multiples of 5 fall from 1.33 to 0.59, a reduction of about
2.3-fold.

Three limits belong with these rows. First, after endpoints are removed no
multiples of 10 survive at $N=5$ or $N=10$ and no interior multiples of 5
survive at $N=5$, so those cells are empty rather than zero. Second, the
surviving sets are small at the shorter ranges: the multiples-of-10 set is
the single integer 10 at $N=20$ and the pair 10 and 20 at $N=30$, so only the
$N=100$ column, with nine and ten members, rests on enough integers to be read
on its own. The same holds for the multiples of 5 that are not multiples of 10:
the set is the single integer 5 at $N=10$, the pair 5 and 15 at $N=20$, and
5, 15 and 25 at $N=30$. Third, every multiple of 5 that is not a multiple of 10 is odd, so
that row overlaps the parity effect reported in the next subsection. Base
checkpoints show a controlled odd-number ratio of 1.01 to 1.16
(Table~\ref{tab:disentangle}), which is lower than the 1.10 to 1.41 seen
here, so parity does not account for the whole of the base figure, but it
accounts for part of it.

Round-number avoidance is not part of the account \citet{kubovy1976}
proposed. Their respondents avoided 77 in the seventies, but the authors
attribute this to 77 looking trite and obvious, not to its being round.
We therefore report round-number avoidance as a property of tuned models,
without a counterpart in the human work we cite. It is the one heuristic in
this section that base checkpoints do not show at all.

\subsection{Odd-number preference, and why controls were needed}

Raw set contrasts make the odd numbers look strongly favoured, and much of
that appearance is arithmetic rather than parity. Odd numbers exclude every
multiple of 10 and half of the multiples of 5, so a model that avoids round
numbers is pushed toward odd values as a by-product. The integer 1 is odd,
sits in several comparison sets, and instruct checkpoints give it almost
exactly zero probability. And when a distribution collapses onto a single
value, a set contrast measures the mode, not the set. Qwen3-4B assigns 17 a
probability of 0.980, and gemma-3-4b-it with prefill assigns it 1.000, so any
such contrast reports little beyond the parity of the mode.

We therefore compare odd against even per member, after removing the multiples
of 5, both endpoints and the modal integer, then renormalising what remains.
Cells with fewer than three members per side are dropped, which removes $N=5$
and $N=10$ entirely. Table~\ref{tab:disentangle} reports the result for 16
base and 33 instruct model-conditions per range (32 for the instruct prime
contrast at $N=30$, where one condition leaves fewer than three non-primes).

\begin{table}[t]
\centering
\caption{Controlled contrasts. The odd-versus-even and prime-versus-non-prime contrasts exclude multiples of five, both endpoints and the modal integer, then renormalise what remains. Endpoint ratios are relative to the average interior integer.}
\label{tab:disentangle}
\begin{tabular}{lrrrrr}
\toprule
Contrast & $N{=}5$ & $N{=}10$ & $N{=}20$ & $N{=}30$ & $N{=}100$ \\
\midrule
\multicolumn{6}{l}{\textbf{base}} \\
\quad odd vs even & -- & -- & 1.09 & 1.01 & 1.16 \\
\quad prime vs non-prime & -- & -- & -- & 1.07 & 1.10 \\
\multicolumn{6}{l}{\textbf{instruct}} \\
\quad odd vs even & -- & -- & 1.64 & 1.24 & 2.09 \\
\quad prime vs non-prime & -- & -- & -- & 2.10 & 1.52 \\
\multicolumn{6}{l}{\textbf{endpoints}} \\
\multicolumn{6}{l}{\textbf{base}} \\
\quad $P(1)$ & 0.600 & 0.406 & 0.362 & 0.321 & 0.356 \\
\quad $P(N)$ & 0.601 & 0.318 & 0.369 & 0.300 & 0.413 \\
\multicolumn{6}{l}{\textbf{instruct}} \\
\quad $P(1)$ & 0.000 & 0.000 & 0.000 & 0.000 & 0.000 \\
\quad $P(N)$ & 0.001 & 0.000 & 0.000 & 0.000 & 0.000 \\
\bottomrule
\end{tabular}
\par\vspace{2pt}\footnotesize A dash means fewer than three members remained on one side of the contrast at that range. Endpoint ratios are rounded to three decimal places; the instruct values at $N=5$ are 0.0001 for $P(1)$ and 0.0007 for $P(N)$. Each cell is the median over 16 base or 33 instruct conditions (upper median for base); the instruct prime contrast at $N=30$ has 32. For the odd-versus-even contrast, Section~\ref{sec:survive} gives the spread and a version with the top two and three integers removed. Values are from the CUDA stack where both stacks ran (Section~\ref{sec:method}).
\end{table}

 \begin{figure}[t]
 \centering
 \includegraphics[width=\linewidth]{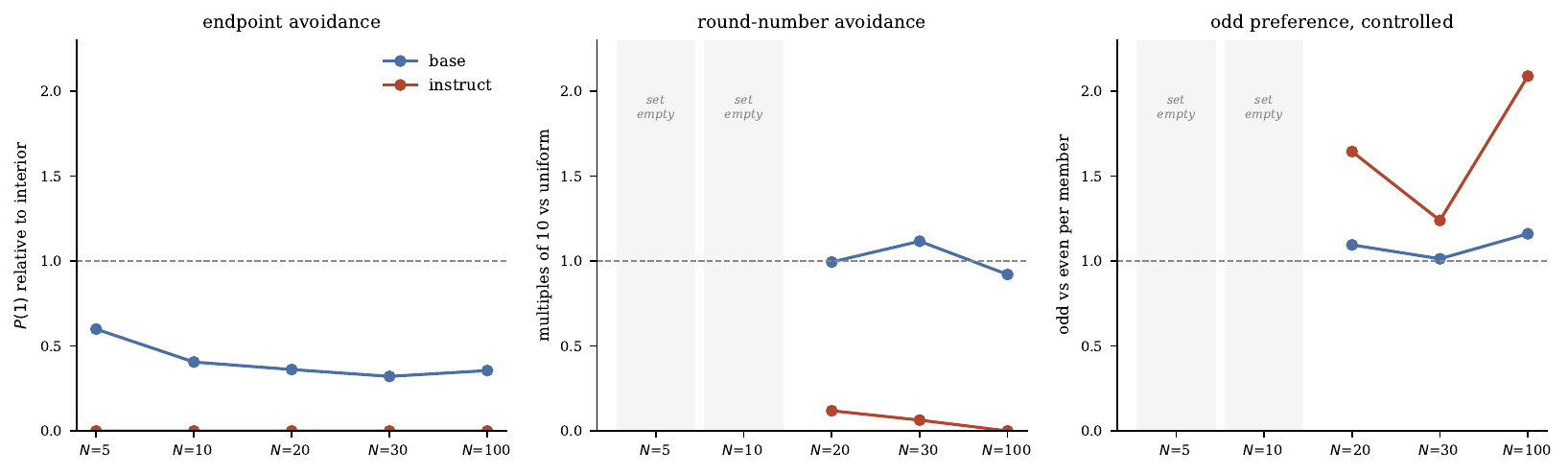}
 \caption{The three heuristics by range. The dashed line at 1.00 is no
   preference. All three panels share a common vertical scale, so the effect
   sizes are directly comparable. Endpoint and round-number ratios fall to
   near zero after instruction tuning; base checkpoints avoid endpoints
   partially and round numbers not at all, sitting at parity in the middle
   panel. The controlled odd preference rises with tuning at the median but is
   a far smaller effect than the other two, and it is shown by a large majority
   of tuned models only at $N=100$. Shaded columns mark ranges where the
   set is empty once endpoints are excluded, which is not the same as a
   ratio of zero.}
 \label{fig:heuristics}
 \end{figure}

The controls flatten the base contrast and leave a tuned one, but the tuned
contrast is consistent across models only at the largest range. Base
checkpoints sit near parity at $N=20$ and $N=30$, at 1.09 and 1.01, with 11
and 9 of 16 conditions above 1. At $N=100$ the base median is 1.16 and all 16
base conditions are above 1 (two-sided sign test, $p = 3.1 \times 10^{-5}$),
so a small odd preference exists before tuning at that range. Instruct medians are 1.64, 1.24 and 2.09 at the three ranges,
but the spread is wide: the interquartile ranges are 0.76 to 3.64, 0.78 to
2.06 and 1.17 to 21.0. At $N=20$, 21 of 33 instruct conditions prefer odd and
12 prefer even, and the same counts hold at $N=30$ (two-sided sign test,
$p=0.16$ at each range). At $N=100$, 27
of 33 prefer odd ($p=0.0003$).

Removing only the modal integer is a weak control when a tuned distribution
puts most of its mass on two or three integers. We therefore repeated the
contrast with the top two and the top three integers removed. Instruct
medians stay above parity, at 1.33, 1.39 and 1.47 with two removed and 1.21,
1.54 and 1.20 with three removed, and 21 to 25 of 33 conditions remain above
1. Base medians fall to parity at $N=20$ and $N=30$ (0.97 to 0.99) and to
1.06 at $N=100$. The odd preference in tuned models is therefore real at the
median but not uniform across models, and only at $N=100$ is it shown by a
large majority of them. It fits the ``oddest digit'' part of
\citet{kubovy1976}'s account, which has respondents avoid even numbers.

The uncontrolled version of this same contrast read 7.32 at $N=5$. That value
had nothing to do with parity: the modal answer at $N=5$ is 3 in 30 of 33
instruct conditions, and 3 is odd. Once the mode is removed, the odd side of
the cell is empty. The same inflation is visible in
Table~\ref{tab:structure}, where instruct odd numbers read 4.40 at $N=5$ and
2.19 at $N=10$ against 1.20 to 1.54 at the longer ranges. The only interior
odd integer at $N=5$ is 3, so that cell reports the mode and nothing else. We quote 7.32 as a caution that, uncontrolled, the headline
odd-number effect measures where the mode falls.

\subsection{Midpoints are essentially not avoided}

The third part of \citet{kubovy1976}'s account is that 5 is avoided as a
traditional midpoint. The models essentially do not show this. At $N=5$ the midpoint, 3,
is the modal answer in 30 of 33 instruct and 13 of 16 base conditions. At the
larger ranges the midpoint itself is often a multiple of 5 (5 at $N=10$, 10 at
$N=20$, 15 at $N=30$, 50 at $N=100$), so its probability would mix midpoint
avoidance with round-number avoidance. We therefore take the middle integer that is not a multiple of 5 (6, 11,
16 and 51 at $N=10$, 20, 30 and 100) and rank its probability against every
other interior integer that is not a multiple of 5. A midpoint that is avoided
would rank low. Its median percentile is 0.83, 0.71, 0.66 and 0.61 for base
checkpoints and 0.67, 0.57, 0.55 and 0.41 for instruct checkpoints. Only for
tuned models at $N=100$ is the midpoint below the middle of its comparison
set, and only slightly. Of the three parts of the human account, the models
share endpoint avoidance clearly, the odd preference at the largest range,
and midpoint avoidance only slightly, for tuned models at $N=100$.

\subsection{What we cannot resolve}

One contrast in Table~\ref{tab:disentangle} cannot be interpreted, and we flag
it rather than bury it: primes. Odd composites that are not multiples of 5 are
overwhelmingly multiples of 3. At $N=30$ the non-prime side of the contrast is
exactly \{9, 21, 27\}, all multiples of 3; at $N=100$, 13 of the 16 non-prime
members are. A model that avoids multiples of 3 and a model that prefers
primes would produce almost the same numbers under this contrast. The second
reading is not arbitrary: \citet{kubovy1976}'s account has respondents avoid
the group 3, 6, 9, that is, multiples of 3.

The medians are 1.07 for base and 2.10 for instruct checkpoints at $N=30$, and
1.10 and 1.52 at $N=100$. We report these values without interpretation: the
prime hypothesis and the multiple-of-3 hypothesis are not separable at these
ranges, and we mark the contrast as unresolved. Settling it requires a range
at which odd composites coprime to 30 are numerous, which means $N=1000$; our
ranges stop at 100.

\section{Range Dependence and the Modal Integer}
\label{sec:range}

Table~\ref{tab:mode-range} gives the modal integer, its frequency, and the median distance from uniform at each of the five ranges.

\begin{table}[!htbp]
\centering
\caption{Modal integer by range. Count is the number of model-conditions whose highest-probability integer equals the mode shown, out of 33 instruct conditions or 16 base conditions per range. Mode/$N$ is the instruct modal integer divided by the range maximum. Median TV is the median total variation distance from uniform across conditions at that range, taken as the upper median: the 17th of 33 sorted values for instruct checkpoints, which is the usual median, and the 9th of 16 for base checkpoints, where the usual median would be 0.166, 0.182, 0.201, 0.187 and 0.234. Values are from the CUDA stack where both stacks ran (Section~\ref{sec:method}).}
\label{tab:mode-range}
\begin{tabular}{lrrrrrrr}
\toprule
 & \multicolumn{4}{c}{Instruct} & \multicolumn{3}{c}{Base} \\
\cmidrule(lr){2-5} \cmidrule(lr){6-8}
Range & Mode & Count & Mode/$N$ & Median TV & Mode & Count & Median TV \\
\midrule
5   & 3  & 30 & 0.60  & 0.680 & 3  & 13 & 0.178 \\
10  & 7  & 28 & 0.70  & 0.736 & 7  & 11 & 0.190 \\
20  & 14 & 12 & 0.70  & 0.704 & 15 & 5  & 0.204 \\
30  & 17 & 14 & 0.567 & 0.739 & 17 & 6  & 0.196 \\
100 & 42 & 13 & 0.42  & 0.842 & 50 & 4  & 0.241 \\
\bottomrule
\end{tabular}
\end{table}

Three of our five ranges coincide with ranges studied by \citet{coronado2025}, who sampled 100 calls per configuration and reported modal answers of 3 and 4 at 1 to 5, 5 and 7 at 1 to 10, and 37, 47 and 73 at 1 to 100. Our exact distributions reproduce the two smaller ranges closely. At $N=5$ the integer 3 is the modal answer in 30 of 33 instruct conditions, and 4 holds the mode in two of the remaining three. At $N=10$, 7 is the modal answer in 28 of 33, with 5 holding the mode in two further conditions. At $N=100$ the agreement is partial. All three of their integers appear as modal answers among the 33 instruct conditions, but rarely: 73 in 6, 47 in 2 and 37 in 1. None of them leads. The leading integer is 42, in 13 of 33 instruct conditions, followed by 73 at 6, and 47 and 53 at 2 each. A plurality mode of 42 at this range has not, to our knowledge, appeared in earlier sampling studies.

\begin{figure}[t]
 \centering
 \includegraphics[width=0.999\linewidth]{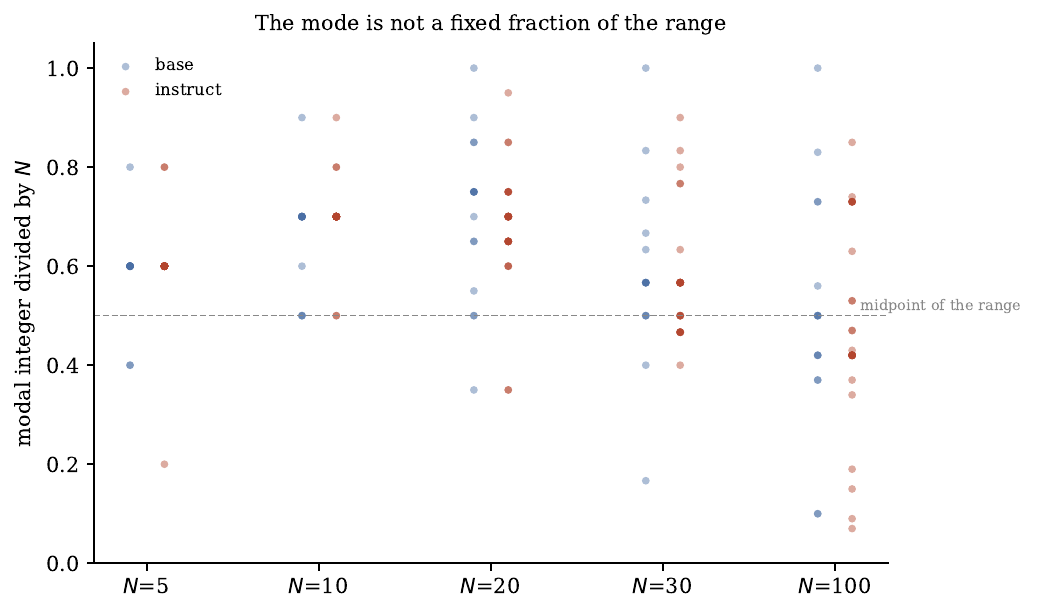}
 \caption{Modal integer divided by the range maximum, one point per
   model-condition, at each of the five ranges. Base checkpoints in blue,
   instruction-tuned in red. The dashed line marks the midpoint. If models
   selected a fixed proportion of the range the points would form a
   horizontal band, and they do not.}
 \label{fig:range}
 \end{figure}

Consensus is high at the two smallest ranges and much lower above them. The leading mode is shared by 91 percent of instruct conditions at $N=5$ (3) and 85 percent at $N=10$ (7), but only by 36 percent at $N=20$ (14), 42 percent at $N=30$ (17) and 39 percent at $N=100$ (42). Base checkpoints are weaker still: 13 of 16 agree on 3 at $N=5$, but only 4 of 16 agree on 50 at $N=100$, where the runners-up are 42 at 3, and 10, 37 and 73 at 2 each. The modal integer is also not a constant fraction of the range. Relative to the range maximum, the leading instruct modes sit at 0.60, 0.70, 0.70, 0.567 and 0.42 for $N=5$, 10, 20, 30 and 100. If models picked a fixed proportion of the range, these five ratios would be equal. They range from 0.42 to 0.70, which is inconsistent with that hypothesis, although five medians cannot rule it out for every model; Figure~\ref{fig:range} shows the spread.

Raw distance from uniform is larger at $N=100$ than at the smaller ranges: in Table~\ref{tab:mode-range} the median instruct TV is 0.680, 0.736, 0.704, 0.739 and 0.842 at the five ranges, and the base median is 0.178, 0.190, 0.204, 0.196 and 0.241. Part of this is arithmetic, because the largest possible TV, $1-1/N$, grows from 0.80 at $N=5$ to 0.99 at $N=100$. Dividing these printed medians by that maximum gives 0.850, 0.818, 0.741, 0.764 and 0.851 for instruct checkpoints and 0.223, 0.211, 0.215, 0.203 and 0.243 for base checkpoints. Neither series rises steadily with the range, and the gap between them stays between 0.53 and 0.63. Concentration is therefore large at every range rather than growing with it. The spread across models is also large: at $N=30$ the interquartile range of instruct TV is 0.53 to 0.92, against 0.11 to 0.30 for base checkpoints.

We turn to 17, the integer most often singled out in discussions of model
randomness. All 15 pairs enter this comparison. $P(17)$ relative to uniform rose after instruction tuning in 10 pairs and fell in 5 (two-sided sign test, $p=0.30$). The median ratio to uniform is 1.71 for base checkpoints and 2.22 for instruct checkpoints. The spread around those medians is enormous. The largest rise is gemma-3-4b, from 0.92 to 29.99 times uniform. The largest falls are Qwen3-1.7B, from 5.34 to 0.01, and Qwen2.5-1.5B, from 4.63 to 0.10. Base checkpoints already sit above uniform at the median, so some elevation of 17 predates tuning, but tuning moves pairs in both directions and can erase the preference entirely. There is no consistent 17 effect in these data.
Table~\ref{tab:seventeen} gives the full breakdown.

\begin{table}[t]
\centering
\caption{Probability assigned to the integer 17 at $N=30$, base against instruction-tuned. Uniform would give 0.0333. The ratio columns give the factor relative to uniform. Values are from the CUDA stack where both stacks ran (Section~\ref{sec:method}).}
\label{tab:seventeen}
\begin{tabular}{lrrrrr}
\toprule
Pair (instruct name) & $P(17)$ base & ratio & $P(17)$ inst. & ratio & change \\
\midrule
google/gemma-3-4b-it & 0.0306 & 0.92 & 0.9996 & 29.99 & 32.62 \\
Qwen/Qwen3-4B & 0.0691 & 2.07 & 0.9798 & 29.39 & 14.19 \\
Qwen/Qwen3-8B & 0.1989 & 5.97 & 0.9569 & 28.71 & 4.81 \\
Qwen/Qwen2.5-7B-Instruct & 0.0426 & 1.28 & 0.4071 & 12.21 & 9.55 \\
meta-llama/Llama-3.2-1B-Instruct & 0.0761 & 2.28 & 0.2331 & 6.99 & 3.06 \\
meta-llama/Llama-3.2-3B-Instruct & 0.0347 & 1.04 & 0.1952 & 5.86 & 5.63 \\
allenai/OLMo-2-1124-7B-Instruct & 0.0381 & 1.14 & 0.1896 & 5.69 & 4.98 \\
meta-llama/Llama-3.1-8B-Instruct & 0.0405 & 1.22 & 0.0739 & 2.22 & 1.82 \\
HuggingFaceTB/SmolLM2-1.7B-Instruct & 0.0569 & 1.71 & 0.0697 & 2.09 & 1.23 \\
Qwen/Qwen2.5-0.5B-Instruct & 0.0448 & 1.35 & 0.0684 & 2.05 & 1.53 \\
Qwen/Qwen3-0.6B & 0.0837 & 2.51 & 0.0353 & 1.06 & 0.42 \\
Qwen/Qwen2.5-3B-Instruct & 0.0934 & 2.80 & 0.0092 & 0.28 & 0.10 \\
Qwen/Qwen2.5-1.5B-Instruct & 0.1544 & 4.63 & 0.0034 & 0.10 & 0.02 \\
Qwen/Qwen3-1.7B & 0.1779 & 5.34 & 0.0004 & 0.01 & 0.00 \\
google/gemma-3-1b-it & 0.0262 & 0.78 & 0.0003 & 0.01 & 0.01 \\
\bottomrule
\end{tabular}
\par\vspace{2pt}\footnotesize The direction is not consistent: $P(17)$ rises in ten pairs and falls in five.
\end{table}

The most useful negative result in this study concerns the parallel with human number choice. \citet{kubovy1976} asked 558 passersby for ``the first number that comes to mind between 0 and 9''; pooled with three earlier studies, 28.4 percent of 1,770 respondents chose 7. They proposed, as a conjecture, that 7 looks spontaneous because it is the ``oddest'' digit: endpoints, the midpoint, even numbers and multiples of 3 are all ruled out. Our tuned models share the endpoint part of that account fully, the odd part at $N=100$, and the midpoint part not at all, and their strongest new effect, round-number avoidance, is not in it. Matching parts of a strategy is also not matching choices. At $N=10$ the tuned models' modal answer, 7, is the human favourite for 0 to 9. Beyond that range we have no human data at matching ranges.
The integer 17 is often described, informally, as the most common ``random'' choice between 1 and 20. At $N=20$ the leading instruct modes are 14, in 12 of 33 conditions, and 13, in 8, followed by 15 in 5, 12 in 3, and 7 and 17 in 2 each. The two conditions in which 17 does lead at $N=20$, SmolLM2-1.7B-Instruct and Phi-4-mini-instruct in the chat condition, give it only 0.133 and 0.206 of the mass. The 17 preference that does appear in our data is specific to $N=30$, where 17 leads in 14 of 33 instruct conditions. Human studies of repeated random generation over small ranges report their own biases, including a preference for small numbers \citep{towse2014}. Parts of the strategy transfer from humans to models; the specific integers do not.

\section{Discussion}

\citet{kubovy1976} argued that human number choice is a search for an answer that looks spontaneous, and proposed, as a conjecture, which answers that rules out: endpoints, the midpoint, even numbers and multiples of 3. Our tuned checkpoints match this account in part. They all but eliminate the endpoints, which base checkpoints already return less often than interior integers. They lean toward odd values, clearly so only at $N=100$. They essentially do not avoid midpoints; the one slight exception is at $N=100$. Their clearest new behaviour, round-number avoidance, has no place in the account and is absent from base checkpoints. A natural hypothesis is that post-training data carries human number-choice habits \citep{kubovy1976, towse2014}. We did not test this hypothesis and cannot: post-training data is not public for any model measured here. Where the specific integers can be compared, they agree at one range only: the tuned models' modal answer at $N=10$ is 7.

The practical implication is simple: do not use a language model as a random number generator. The sharper point is that the failure is structured, not noisy. When the choice is over integers, a tuned model asked to choose arbitrarily systematically avoids the first option, the last option and round values. Whether the same holds for choices over non-numeric options is untested here. This plausibly affects any task in which a model picks arbitrarily among options, such as tie-breaking, assigning items to conditions, or selecting among equivalent candidates. We have not tested those tasks, so this remains a conjecture. Earlier sampling-based work reached compatible conclusions: models fail to produce uniform output or to reproduce requested distributions \citep{hopkins2023}, behave poorly as dice \citep{zhao2026}, and base models sit closer to uniform than aligned ones \citep{westpotts2025}, which our exact measurements confirm.

The remedy is mechanical. If the model's distribution can be inspected, renormalise the logits over the valid candidates and sample from the renormalised distribution. If it cannot, as with closed models, use an external generator and pass the result to the model. This is a one-line remedy rather than a research contribution, and we present it as such.

Exact enumeration removes sampling error from the primary measurement, but it also removes a check. A tokenisation error, for example, produces plausible-looking numbers with no visible symptom. Sampling validation should therefore be run for at least one condition per model as a routine companion to enumeration. We did not meet that standard: we validated one condition on one model, so the recommendation is drawn from the method rather than from our own coverage. Our cross-stack replication confirms numerical agreement between two inference stacks; it would not catch an error shared by both, such as a mistake in the enumeration logic itself.

\section*{Limitations}

\begin{itemize}
\item Open-weight models only, and none larger than 8B parameters. Closed models do not expose their distributions, and the practically important behaviour is theirs.
\item One validation case. Sampling validation covers one model in one condition.
\item Non-standard spellings are not scored. $P(n)$ covers the bare and leading-space spelling of each integer in the tokeniser's own encoding; spellings such as ``1'' followed by ``7'' where ``17'' is one token are omitted, and their mass was not measured.
\item Three Gemma 2 9B runs were not completed, because the disk was full:
  the base checkpoint under the completion prompt, and the instruct
  checkpoint under both the chat and prefill conditions.
\item Seven model lines ran on the Metal stack only, and three Gemma 3 checkpoints gave valid output only on the CUDA stack, so those conditions are not replicated.
\item Qwen3 is measured with its reasoning block closed, which is not its default.
\item One prompt is primary. Prompt variants were not run at scale.
\item English prompts only. \citet{coronado2025} varied language and found effects.
\item Quantisation was not varied. All measurements are bfloat16 or float16.
\item Statistics are descriptive. The primary quantities are exact, and comparisons across models use only sign tests, which assume models are independent.
\item Base and instruct conditions use different prompt formats by necessity. We control the answer position but cannot make the formats identical.
\item Small sets at small ranges. The round-number sets have one to three members at $N \le 30$, so only the $N=100$ column rests on many integers.
\item The prime hypothesis is confounded with multiples of 3 and remains unresolved. At $N=30$ the non-prime comparison set is exactly $\{9, 21, 27\}$, all multiples of 3.
\item Model size effects are not reported. Parameter count is confounded with tokeniser, corpus, architecture and post-training recipe.
\end{itemize}

\section{Conclusion}

Instruction tuning does not merely make integer generation non-uniform. Across all 15 usable pairs it moves the distribution away from uniform, installs one heuristic and sharpens another. Round numbers are avoided only after tuning. Endpoints, already returned at 0.30 to 0.60 of the interior rate by base checkpoints, are almost never returned after it. A preference for odd values appears in tuned models at the median, but only at $N=100$ is it shown by most of them, and base checkpoints show a weak version there too. Endpoint avoidance and the odd preference fit the account \citet{kubovy1976} proposed for human choice; round-number avoidance is not part of that account, the midpoint avoidance it predicts is essentially absent (one slight exception, for tuned models at $N=100$), and the specific integers chosen do not match human data beyond 7 at $N=10$. The target of the concentration is model-specific rather than shared and is not predictable from the base checkpoint. It is a property of the model's own distribution rather than an artefact of sampling settings: at temperature 1.0, nucleus truncation adds a median of 0.035 and at most 0.082 to the distance from uniform. The one large sampler effect we found occurs at temperature 0.7 and differs between stacks (Section~\ref{sec:concentration}).

\section*{Reproducibility}

We release the exact probability distributions for all 85 completed runs, the measurement code, the runner, and the analysis scripts. These are sufficient to reproduce every table and figure in the paper. The repository is available at \url{https://zenodo.org/records/23674051}.  

\appendix

\section{Excluded Rows}
\label{app:exclusions}

Two exclusion thresholds are used in this paper. Both were fixed before the
analysis was run. This appendix accounts for every row they remove, individually where the
count is small and by group otherwise, so that nothing
is discarded silently.

\subsection{The primary threshold}

A row is excluded if its valid\_mass falls below $10^{-4}$. Below that value
the conditional distribution $P_{\mathrm{cond}}(n) = P(n) / \text{valid\_mass}$
is a ratio of near-zero quantities and is numerically meaningless: the
reported shape is determined by floating-point residue rather than by
anything the model does. This rule removes 1,247 of 12,325 rows, or 10.1
percent, leaving 11,078.

Rows removed by this rule include four groups. The first is greedy
decoding on models whose single most likely first token is a preamble word
rather than a digit. For Qwen3-0.6B in the chat condition with the reasoning block closed, the
first generated token is ``Sure'' with probability 0.407, followed by ``Let''
at 0.192 and ``I'' at 0.071. With the block open the first token is
\texttt{<think>} instead, as reported in Section~\ref{sec:method}. The probability of a complete bare integer is
therefore zero under greedy decoding, and the metrics are undefined. The
second is reasoning-model conditions measured with the reasoning block
open, where valid\_mass falls to around $6 \times 10^{-15}$. These are
excluded from all reported analyses; the thinking-off condition is used
throughout instead, as described in Section~\ref{sec:method}. The third is the
whole Mistral-7B-Instruct-v0.3 prefill run. Its first digit is highly
probable (for example 0.797 for ``3'' at $N=5$), but the model almost never
stops after the integer, so valid\_mass is below $5 \times 10^{-5}$ in
every row. This is why no Mistral pair appears in
Table~\ref{tab:base_vs_instruct}. The fourth is the Metal-stack runs of
gemma-3-4b-pt, gemma-3-4b-it (both conditions) and gemma-3-1b-pt, which
returned no valid probabilities at any range or setting. We did not diagnose the cause. The same checkpoints are valid on the CUDA stack, and
those rows are used.

\subsection{The set-membership threshold}

For every analysis of the model prior, including the paired comparison and
the analyses that aggregate probability mass over subsets of the integers, a
stricter threshold of valid\_mass $\geq 0.10$ is applied. The reason differs
from the one behind the primary threshold. A row with valid\_mass of 0.03 is
numerically well conditioned, but its conditional distribution describes what
the model does on the three percent of occasions when it follows the
instruction, not what it does typically. Reporting such a row alongside a row
with valid\_mass near 1.000 would compare two different things.

At $N=30$ under the model prior, this threshold removes exactly three
model-conditions. They are listed in full in Table~\ref{tab:excluded}.

\begin{table}[h]
\centering
\caption{Every model-condition excluded from the $N=30$ headline analysis by
the valid\_mass $\geq 0.10$ rule. All three are reported here rather than
omitted. Note that the first row would otherwise appear as the most uniform
distribution in the paper.}
\label{tab:excluded}
\begin{tabular}{lrrrrr}
\toprule
Model [condition] & TV & Mode & $p$(mode) & Valid mass & Ent.\ def. \\
\midrule
mistralai/Mistral-7B-v0.3 [base]   & 0.061 & 7  & 0.047 & 0.027 & 0.012 \\
google/gemma-2-2b-it [prefill]     & 0.542 & 17 & 0.357 & 0.026 & 1.015 \\
google/gemma-2-2b-it [chat]        & 0.573 & 17 & 0.361 & 0.002 & 1.109 \\
\bottomrule
\end{tabular}
\end{table}

The first row deserves comment, because excluding it changes a headline
number. Mistral-7B-v0.3 base has a total variation distance from uniform of
0.061, which is lower than any retained model-condition in the paper. Read
naively it is the most uniform model we measured. But its valid\_mass is
0.027, so that distribution describes the 2.7 percent of its output that is
a bare terminated integer. With the row excluded, the most uniform retained
model-condition is OLMo-2-7B base at 0.063, whose valid\_mass is 0.995.

The same three model-conditions fail this threshold at every range, so every
model-prior analysis uses 16 base and 33 instruct conditions at each range.
At every range these three have valid\_mass of 0.027 or less. The lowest
retained valid\_mass at each range belongs to Qwen3-0.6B in the chat
condition: 0.253, 0.112, 0.273, 0.211 and 0.163 at $N=5$, 10, 20, 30 and 100.
Because the gemma-2-2b-it rows are excluded, the paired comparison and the
$P(17)$ comparison both cover 15 pairs. For the record, the excluded
gemma-2-2b pair moves in the same direction as the others, from a total
variation distance of 0.106 to 0.542.

\subsection{Runs not completed}

Three runs were not completed, because the machine ran out of storage
partway through the schedule (402 MB free against shards of about 5 GB): google/gemma-2-9b under the completion prompt,
and google/gemma-2-9b-it under both the chat and prefill conditions. No
partial data from these runs is included anywhere. Gemma 2 is represented in
the paper by its 2B checkpoints and Gemma 3 by its 1B and 4B checkpoints, so
the model line is covered, but the 9B size point is missing.

\section{Models and Stacks}
\label{app:models}

Table~\ref{tab:models} lists every checkpoint attempted. The 18 model lines are
the 17 base and instruction-tuned lines plus Phi-4-mini-instruct, which has no
public base checkpoint. Of the 17, 15 enter the paired comparison: Mistral-7B
is lost to the primary threshold and gemma-2-2b to the set-membership
threshold. Each condition was run at all five ranges and all 29 decoding
configurations, giving 145 result rows per run. The Metal stack completed 53
runs and the CUDA stack 32, for 85 runs and 12,325 rows.

\begin{table}[h]
\centering
\small
\caption{Every checkpoint attempted. Stacks lists the stacks that gave valid output; the Note column gives exceptions by condition. Base checkpoints ran in the completion condition, instruction-tuned checkpoints in the chat and prefill conditions. Stacks: CUDA (RTX 4070 Ti, bfloat16) and Metal (Mac Mini, float16). Model names omit the vendor prefix (Qwen, google, meta-llama, allenai, HuggingFaceTB, mistralai, microsoft).}
\label{tab:models}
\resizebox{\linewidth}{!}{%
\begin{tabular}{lllll}
\toprule
Line & Base checkpoint & Instruct checkpoint & Stacks & Note \\
\midrule
Qwen2.5 & Qwen2.5-0.5B & Qwen2.5-0.5B-Instruct & both &  \\
 & Qwen2.5-1.5B & Qwen2.5-1.5B-Instruct & both &  \\
 & Qwen2.5-3B & Qwen2.5-3B-Instruct & both &  \\
 & Qwen2.5-7B & Qwen2.5-7B-Instruct & Metal &  \\
Qwen3 & Qwen3-0.6B-Base & Qwen3-0.6B & both & thinking off \\
 & Qwen3-1.7B-Base & Qwen3-1.7B & both & thinking off \\
 & Qwen3-4B-Base & Qwen3-4B & both & thinking off \\
 & Qwen3-8B-Base & Qwen3-8B & Metal & thinking off \\
Gemma 2 & gemma-2-2b & gemma-2-2b-it & both & tuned fails 0.10 \\
 & gemma-2-9b & gemma-2-9b-it & none & not completed \\
Gemma 3 & gemma-3-1b-pt & gemma-3-1b-it & both & base: Metal run invalid \\
 & gemma-3-4b-pt & gemma-3-4b-it & CUDA & Metal runs invalid \\
Llama 3 & Llama-3.2-1B & Llama-3.2-1B-Instruct & Metal &  \\
 & Llama-3.2-3B & Llama-3.2-3B-Instruct & Metal &  \\
 & Llama-3.1-8B & Llama-3.1-8B-Instruct & Metal &  \\
OLMo 2 & OLMo-2-1124-7B & OLMo-2-1124-7B-Instruct & Metal &  \\
SmolLM2 & SmolLM2-1.7B & SmolLM2-1.7B-Instruct & both &  \\
Mistral & Mistral-7B-v0.3 & Mistral-7B-Instruct-v0.3 & Metal & no pair; see A.1 \\
Phi-4 & none & Phi-4-mini-instruct & both & instruct only \\
\bottomrule
\end{tabular}}
\end{table}

\clearpage

\bibliographystyle{plainnat}
\bibliography{refs_2026-10-15_v6}

\end{document}